%% file: main.tex
\documentclass[10pt,twocolumn,letterpaper]{article}

\usepackage{cvpr}              % To produce the CAMERA-READY version
\usepackage{algorithm} 
\usepackage{algpseudocode}

\usepackage{graphicx}
\usepackage{pifont}
\usepackage{multirow}
\usepackage{xcolor}
\usepackage{adjustbox}

\definecolor{darkgreen}{RGB}{1,50,32}
\definecolor{forestgreen}{RGB}{34,139,34}
\definecolor{cvprblue}{rgb}{0.21,0.49,0.74}
\usepackage[pagebackref,breaklinks,colorlinks,allcolors=cvprblue]{hyperref}

\def\paperID{32} % *** Enter the Paper ID here
\def\confName{CVPR}
\def\confYear{2025}

\title{SAGA: Semantic-Aware Gray color Augmentation for Visible-to-Thermal Domain Adaptation across Multi-View Drone and Ground-Based Vision Systems}

\author{Manjunath D$\ ^{1}$, Aniruddh Sikdar$\ ^{2}$ , Prajwal Gurunath$\ ^{1}$, Sumanth Udupa$\ ^{1}$, Suresh Sundaram$^{1,2}$        
\\  
$^{1}$Department of Aerospace Engineering, Indian Institute of Science, Bengaluru, India\\
$^{2}$Robert Bosch Centre for Cyber Physical Systems, Indian Institute of Science, Bengaluru, India\\
{\tt\small \{manjunathd1,aniruddhss, vssuresh\} @iisc.ac.in}  
}
\begin{document}
\maketitle
\input{sec/0_abstract}

\input{sec/1_intro}

\input{sec/2_relatedworks}
\input{sec/4_methodology}
\input{sec/5_experimental}
\input{sec/6_abalation}
\input{sec/7_conclusion}
{
    \small
    \bibliographystyle{ieeenat_fullname}
    \bibliography{main}
}


\end{document}

%% file: sec/0_abstract.tex
% \begin{abstract}
% The ABSTRACT is to be in fully justified italicized text, at the top of the left-hand column, below the author and affiliation information.
% Use the word ``Abstract'' as the title, in 12-point Times, boldface type, centered relative to the column, initially capitalized.
% The abstract is to be in 10-point, single-spaced type.
% Leave two blank lines after the Abstract, then begin the main text.
% Look at previous \confName abstracts to get a feel for style and length.
% \end{abstract}

\begin{abstract}

Domain-adaptive thermal object detection plays a key role in facilitating visible (RGB)-to-thermal (IR)  adaptation by reducing the need for co-registered image pairs and minimizing reliance on large annotated IR datasets. However, inherent limitations of IR images, such as the lack of color and texture cues, pose challenges for RGB-trained models, leading to increased false positives and poor-quality pseudo-labels. To address this, we propose Semantic-Aware Gray color Augmentation (SAGA), a novel strategy for mitigating color bias and bridging the domain gap by extracting object-level features relevant to IR images. Additionally, to validate the proposed SAGA for drone imagery, we introduce the IndraEye, a multi-sensor (RGB-IR) dataset designed for diverse applications. The dataset contains 5,612 images with 145,666 instances, captured from diverse angles, altitudes, backgrounds, and times of day, offering valuable opportunities for multimodal learning, domain adaptation for object detection and segmentation, and exploration of sensor-specific strengths and weaknesses. IndraEye aims to enhance the development of more robust and accurate aerial perception systems, especially in challenging environments. Experimental results show that SAGA significantly improves RGB-to-IR adaptation for autonomous driving and IndraEye dataset, achieving consistent performance gains of +0.4$\%$ to +7.6$\%$ (mAP) when integrated with state-of-the-art domain adaptation techniques. The dataset and codes are available at \href{https://github.com/airl-iisc/IndraEye}{https://github.com/airl-iisc/IndraEye}
% As multimodal deep learning methods have shown promising results, it is relevant to take advantage of such algorithms to deal with EO-IR images to achieve superior performance in harsh environmental conditions. 
%Aerial object detection and segmentation are very challenging tasks due to factors such as slant angle, occlusion, scale variations and domain shift, also there are only a few datasets that deal with such problems. A good multimodal sensing using RGB, and Thermal data requires an accurate co-registration technique, this can pose a significant challenge since the sensors used for each modality may have varying field of views (FOV), resolution and sensing capabilities. Considering these challenges we propose a novel multi-sensor i.e. RGB-IR and unco-registered dataset with mostly same background and objects designed for object detection and segmentation. 
%Effective RGB-to-IR domain adaptation requires addressing both the inherent limitations of IR imaging and the algorithmic biases of RGB-trained models
% to obtain high accuracy on IR images using EO images through knowledge distillation.
    % We believe that this dataset can help in understanding the two modalities in hard conditions on aerial and ground vehicles. 
%The dataset and codes are available at \href{https://bit.ly/indraeye}{https://bit.ly/indraeye}
\end{abstract}

%% file: sec/1_intro.tex
\section{Introduction}
\label{sec:intro}
%However, since IR cameras have less semantic information, it needs to be learned together.
% The synergistic relationship between RGB and thermal (RGB-T) imaging modalities stems from their orthogonal physical sensing principles, creating a robust perceptual system that overcomes the limitations of unimodal approaches. This complementary nature manifests across five key dimensions

Deep learning has advanced real-world applications like autonomous driving \cite{udupa2024mrfp} and surveillance \cite{sikdar2023deepmao}. Robust object detection for aerial perception is essential for Unmanned Aerial Vehicles (UAVs) to operate effectively in challenging, low-light environments, facilitating critical tasks such as infrastructure inspections \cite{senthilnath2024metacognitive}, environmental monitoring \cite{john2024efficient} and surveillance \cite{sarker2024transformer, sikdar2022fully}. Most vision models focus on visible-spectrum (RGB) cameras due to the availability of extensive datasets, however, their robustness diminishes in low-light conditions \cite{ha2017mfnet, sikdar2023contrastive}. Increasing research in robotics and computer vision highlights thermal infrared (IR) imaging for its effectiveness in harsh weather and low visibility, as IR cameras capture unique spectral data and penetrate dust and smoke. While IR cameras are valuable, their images generally contain less semantic information than RGB images, leading to performance drops in deep learning models for downstream tasks \cite{sikdar2024skd, Sikdar_Teotia_Sundaram_2025}.  
%Integrating RGB and IR data mitigates these limitations, enhancing environmental perception by combining the strengths of both modalities. This fusion benefits applications such as nighttime surveillance, search and rescue \cite{john2024efficient},  and all-weather autonomous driving systems \cite{liu2023multi}.
\begin{figure*}[t]
    \centering
    \includegraphics[scale=0.35]{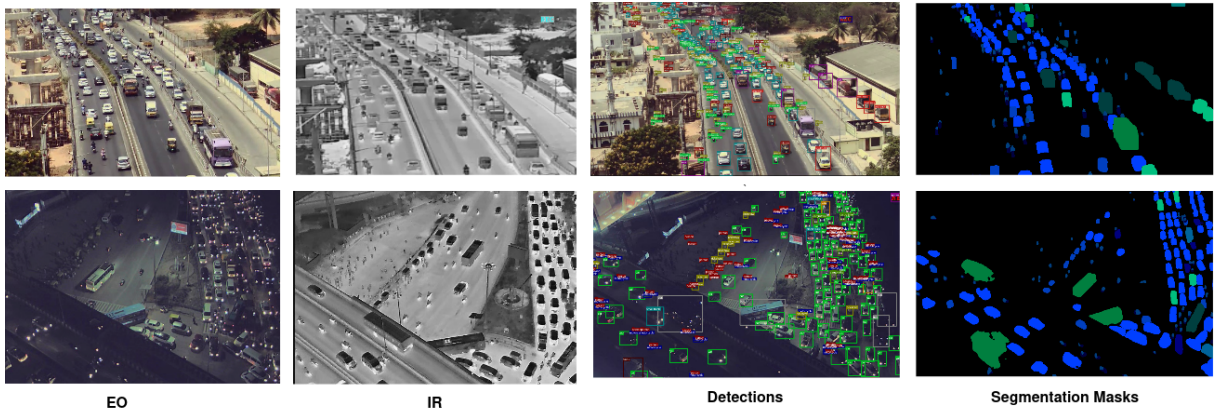}
    \setlength{\abovecaptionskip}{-2pt} 
    \setlength{\belowcaptionskip}{-11pt}
    \caption{  Snapshots from the IndraEye dataset  showing different modalities RGB, IR and complete semantic annotations for detection \& segmentation tasks taken from different slant angles}
    \label{fig:inra_NEW}
\end{figure*}
With affordable IR sensors, deep multimodal fusion uses RGB-IR integration to outperform unimodal methods \cite{valada2020self, kutuk2022semantic}. However, multimodal techniques face three key challenges: \textit{(i) Limited Annotated Thermal Datasets:} The scarcity of labeled IR data limits training high-performance detection models. \textit{(ii) Dependency on Co-Registered RGB-IR Image Pairs:} Accurate fusion relies on precisely aligned RGB and IR images, but achieving co-registration is challenging due to hardware constraints and sensor-specific variations. These challenges frequently result in alignment errors, which can substantially impact performance, especially in drone applications \cite{brenner2023rgb, song2024misaligned}, and \textit{(iii) Domain Shift Between Modalities:} IR images differ significantly from RGB, causing domain shift and reducing model performance on IR data. Conventional Unsupervised Domain Adaptation (UDA) for object detection has been predominantly studied in the context of RGB images. In contrast, domain-adaptive thermal object detection \cite{do2024d3t} aims to enhance detection performance in the infrared (IR) spectrum by leveraging UDA techniques. Furthermore, UDA can be extended for RGB-to-IR domain adaptation to address the inherent challenges of deep multimodal fusion models. By minimizing the domain discrepancy between the source (i.e., RGB images) and target (i.e., IR images) domains, it alleviates the constraints of co-registered image pairs and mitigates the reliance on extensive annotated IR datasets, thereby enhancing the adaptability of object detection models across modalities.

Domain-adaptive thermal object detectors use the Mean Teacher (MT) framework \cite{tarvainen2017mean}, where the teacher model generates pseudo-labels from the target domain to guide the student. The teacher’s weights are updated via an exponential moving average of the student’s weights for improved cross-domain adaptation. Object detection requires fine-grained, localized representations, but RGB-trained models often over-rely on color cues and texture patterns, which are absent in monochromatic IR images \cite{zhang2022fmcnet, ustun2023spectral}. Additionally, IR imaging struggles to capture fine edges visible in RGB due to reduced edge clarity \cite{tennant2011limits, ma2024aerialirgan}. These inherent limitations in color and structural representation pose challenges for domain adaptation, leading to poor-quality pseudo-labels. Additionally, domain-adaptive thermal object detection for drone imagery is still in its early stages and presents significant challenges in the RGB-IR domain due to the distinct nature of aerial perspectives. The adaptation process is further challenged by dynamic environmental conditions, diverse viewing angles, and factors like weather fluctuations, and inconsistencies in acquisition parameters. Perhaps the most critical limitation is the lack of large-scale annotated datasets, which impedes the advancement of robust adaptation techniques for drone-based applications \cite{zeng2024unsupervised, lee2024dehazing}.

To address this challenge, we introduce Semantic-Aware Gray color Augmentation (SAGA), a novel augmentation strategy applied to RGB images to mitigate color bias and bridge the domain gap between RGB and IR. The domain gap often results in poor-quality pseudo-labels from the teacher model \cite{wu2017rgb}, which SAGA alleviates by extracting object-level features more relevant to IR images. SAGA selectively converts only object instances to grayscale while preserving the background’s color information, creating a hybrid RGB-grayscale image that maintains the structural integrity of objects. As a simple instance-level augmentation strategy, SAGA is applied exclusively to the source domain using RGB images. By training the teacher model to perceive object instances in a manner similar to their IR counterparts, SAGA facilitates precise adaptation of localized features in the target domain, consistently enhancing performance on IR imagery in a domain adaptation setting.

To validate SAGA for RGB-IR drone imagery, we introduce IndraEye, a comprehensive dataset tailored for aerial perception.  It includes diverse object categories like road vehicles and pedestrians, captured across varied angles, backgrounds, and scales, spanning urban areas to highways with different population densities and environmental conditions. On average, each RGB image contains 35 annotated instances, with a strong emphasis on dense traffic scenarios. Figure \ref{fig:inra_NEW} showcases examples of RGB and IR images captured from different slant angles. IndraEye supports multitask learning, including object detection and semantic segmentation, making it a crucial benchmark for evaluating and improving model robustness across multiple modalities and tasks. Additionally, we introduce a structured benchmark for RGB-to-IR domain adaptation, making it one of the first for drone imagery . The main contributions of this paper are:
% \footnote{IndraEye will be publicly available for the vision community, along with systematic benchmarking for various tasks.}

\begin{enumerate}
    % \item We propose Semantic-Aware Gray color Augmentation (SAGA), a novel instance-level augmentation strategy. SAGA creates hybrid RGB-grayscale images that preserve object structure, helping to mitigate color bias and narrow the domain gap for RGB-to-IR domain adaptation.
    \item We propose Semantic-Aware Gray color Augmentation (SAGA), a novel instance-level augmentation strategy. SAGA creates hybrid RGB-grayscale images that preserve object structure, mitigating color bias and narrowing the domain gap for RGB-to-IR domain adaptation.
    \item  We introduce the IndraEye dataset, that includes diverse scenes captured from seven different locations with varying slant angles and heights, resulting in significant object scale variations. The dataset includes 145,666 dense instances across 13 classes, covering various road vehicles and pedestrians in both RGB and IR modalities, supporting object detection and segmentation.
    \item SAGA enhances RGB-to-IR adaptation for autonomous driving and IndraEye drone imagery. Integrated with state-of-the-art domain adaptation techniques, it achieves consistent performance gains of +0.4$\%$ to +7.6$\%$  (mAP) across different scenarios.

\end{enumerate}

\begin{table*}[h]
\centering
\label{dataset_desc}
\fontsize{9pt}{9pt}\selectfont
\begin{tabular}{c c c c c c c c}
\toprule
\textbf{Scene} & \textbf{\begin{tabular}[c]{@{}c@{}}Altitude\\ (metres)\end{tabular}} & \textbf{\begin{tabular}[c]{@{}c@{}}Dynamic Angle \\ range (degrees)\end{tabular}} & \textbf{\begin{tabular}[c]{@{}c@{}}RGB Instances\\ daytime\end{tabular}} & \textbf{\begin{tabular}[c]{@{}c@{}}RGB Instances\\ nightime\end{tabular}} & \textbf{\begin{tabular}[c]{@{}c@{}}IR Instances\\ daytime\end{tabular}} & \textbf{\begin{tabular}[c]{@{}c@{}}IR Instances\\ nightime\end{tabular}} & \textbf{Scale-variablity} \\
\midrule
A   & 30                                                                   & 10-25                                                                   & 1982                                                                    & -                                                                        & 9247                                                                    & -                                                                        & Mid                       \\ 
B   & 30                                                                   & 10-25                                                                   & 14149                                                                   & -                                                                        & 8215                                                                    & -                                                                        & Mid                       \\ 
C         & 60                                                                   & 5-50                                                                    & 37752                                                                   & 5312                                                                     & 7559                                                                    & 1285                                                                     & High                      \\ 
D         & 12                                                                   & 20-40                                                                   & 5369                                                                    & -                                                                        & -                                                                       & -                                                                        & Low                       \\ 
E        & 12                                                                   & 20-40                                                                   & 5394                                                                    & 3774                                                                     & -                                                                       & 6124                                                                     & Low                       \\ 
F         & 12                                                                   & 10-30                                                                   & 4728                                                                    & -                                                                        & 8759                                                                    & -                                                                        & Mid                       \\ 
G            & 7                                                                    & 20-40                                                                   & 2234                                                                    & -                                                                        & 1936                                                                    & 2024                                                                     & Low                       \\ 
H            & 7                                                                    & 10-30                                                                   & 3033                                                                    & 2971                                                                     & 3731                                                                    & 7413                                                                     & High                      \\  
\bottomrule
\end{tabular}
\caption{IndraEye dataset description: with altitude of imaged scene, dynamic angle ranges, scene-wise instances, and scale-variability of each scene.}
\label{table: dataset description_class}
\end{table*}

%% file: sec/2_relatedworks.tex
\section{Related works}
\label{sec:formatting}
\begin{table*}[ht]

\centering
\label{table:Dataset comparison}
\begin{adjustbox}{max width=1.0\textwidth}
\begin{tabular}{l c c c c c c c}
\toprule
\textbf{Datasets} & \multicolumn{1}{l}{\textbf{Multi-sensory}} & \multicolumn{1}{l}{\textbf{Diverse Viewpoints}} & \multicolumn{1}{l}{\textbf{Diverse backgrounds}} & \multicolumn{1}{l}{\textbf{Diverse classes}} & \multicolumn{1}{l}{\textbf{Diverse illumination}} & \multicolumn{1}{l}{\textbf{Detection}} & \multicolumn{1}{l}{\textbf{Segmentation}}\\
\midrule
DOTA \cite{xia2018dota}             & \ding{55}                                          & \ding{55}                                                & \ding{55}                                                 & \checkmark                                             & \ding{55}                                    & \checkmark                 & \ding{55}          \\
HIT-UAV \cite{suo2023hit}         & \ding{55}                                           & \checkmark                                                & \checkmark                                                 & \ding{55}                                             & \ding{55}                             & \checkmark                   & \ding{55}               \\
VisDrone \cite{cao2021visdrone}        & \ding{55}                                           & \checkmark                                                & \checkmark                                                 & \checkmark                                             & \checkmark                        & \checkmark                   & \ding{55}                                       \\
UAVDT  \cite{du2018unmanned}           & \ding{55}                                           & \checkmark                                                & \checkmark                                                 & \checkmark                                             & \checkmark                        & \checkmark                   & \ding{55}                                       \\
Vedai \cite{razakarivony2016vehicle}            & \checkmark                                           & \ding{55}                                                & \checkmark                                                 & \checkmark                                             & \checkmark               & \checkmark                   & \ding{55}                                                \\
M3FD \cite{liu2022target}             & \checkmark                                           & \ding{55}                                                & \checkmark                                                 & \checkmark                                             & \checkmark                        & \checkmark                   & \ding{55}                                       \\
FLIR \cite{fliresh}            & \checkmark                                           & \ding{55}                                                & \checkmark                                                 & \checkmark                                             & \checkmark                               & \checkmark                   & \ding{55}                                \\
MSRS \cite{tang2022piafusion}          & \checkmark                                           & \ding{55}                                              & \checkmark                                                 & \checkmark                                             & \checkmark                                & \ding{55}                   & \checkmark                               \\
InfraParis \cite{franchi2024infraparis}  & \checkmark                                           & \ding{55}                                              & \checkmark                                                 & \checkmark                                             & \checkmark                                & \checkmark                   & \checkmark                               \\
{\textbf{IndraEye (Ours)}}   & \checkmark                                           & \checkmark                                                & \checkmark                                                 & \checkmark                                             & \checkmark                                & \checkmark                   & \checkmark                               \\
\bottomrule
\end{tabular}
\end{adjustbox}
\caption{Qualitative comparison of multiple aerial vehicle object detection datasets.}
\label{Qual_comp}
\end{table*}

% Recent advancements in object detection using aerial imagery have been significantly influenced by the availability of diverse datasets captured through Electro-Optical (EO) cameras. These datasets are generally classified based on their viewing angles: (1) nadir or bird’s-eye view datasets, such as VEDAI and DOTA, and (2) oblique or slant-angle datasets, including VisDrone and HIT-UAV. Prominent contributions from datasets like UAVDT and VisDrone have underscored the critical role of UAV-based object detection, driving considerable progress in this domain. These datasets are widely utilized for applications such as aerial perception and traffic monitoring, offering images with diverse viewpoints, backgrounds, and lighting conditions. However, object detection algorithms often face challenges in nighttime settings. While EO cameras deliver high-resolution imagery with detailed textures under adequate lighting, their performance declines in low-light environments due to the lack of sufficient texture details, resulting in reduced detection accuracy.
\textbf{Thermal Object Detection}
% With the technical advances in computing and sensors, the work on RGB object detection has gained significant progress in the fields of robotics, autonomous vehicles for various applications. Similar trends are observed in IR object detection recently. 
% \vspace{-1em}
%%%%%%%%%%%%%%%%CORRECT%%%%%%%%%%%%%%%%%%%%%%%%
% Object detection task using thermal images has been regarded as important area of research in computer vision mainly for military applications such as surveillance, drone reconnaissance, and autonomous driving. Particularly, Aerial Thermal Object Detection (ATOD) faces lot of challenges due to variations in size of the objects and background.  In order to improve the thermal object detection works such as \textbf{cite some gen ai stuff} use GAN to generate IR or RGB images or vice-versa using one of the modality to avoid manual dataset annotation. Several works including \textbf{cite fusion algos} have been proposed to use both modalities by fusing the images when the dataset contains images that are co-registered. Recently few works make use of misaligned RGB-IR images for object detection as seen in \textbf{cite the IEEE transaction and science direct paper}. This clearly shows that the trend of relying on co-registerd dataset is not practically feasible and the need to account for this problem using modern approaches such as Domain adaptation and feature fusion. Several image fusion methods \cite{liu2022target}, \cite{zhao2023cddfuse} have also been proposed to integrate co-registered modalities.
%%%%%%%%%%%%%%%%%%%%% %%%%%%%%%%%%%%%%%%%%%%%%%%%%%%%%%%%%%
Object detection using thermal images has become a critical research area in computer vision, with Aerial Thermal Object Detection (ATOD) posing significant challenges due to variations in object sizes and backgrounds \cite{ding2021object}. To address these challenges, GANs have been explored \cite{devaguptapu2019borrow, sikdar2024ssl} for IR generation to reduce manual annotation and enhance object detection performance. Recent studies \cite{song2024misaligned, ma2024hierarchical} have explored object detection with misaligned RGB-IR images, revealing the limitations of co-registered datasets. These works emphasize the necessity of advanced techniques such as domain adaptation to address this challenge effectively. \\
% Despite the advancements in RGB cameras, they suffer a major set-back during uncertain environmental conditions such as low-illumination, fog, rain, etc. Due to which a lot of critical applications in context of surveillance and military missions use thermal camera along with RGB to tackle such cases \textbf{cite aniruddhs work}. These works mainly involve sensor fusion on the image space \textbf{cite sensor fusion algos} which require both RGB and thermal images to be co-registered, and they do not account for the pixel shifts that occur during registration process. For which recently D3T \cite{do2024d3t} has proposed domain adaptation technique to handle RGB and IR modalities simultaneously. This work follows the traditional mean-teacher architecture that was originally proposed for Unsupervised Domain adaptation \textbf{Cite mean-teacher work}. Several algorithms such as \cite{cao2023contrastive}, 
\textbf{Domain Adaptive Thermal Object Detection}
UDA for RGB-to-IR adaptation reduces domain gaps between a labeled source and an unlabeled target. It minimizes discrepancies by aligning features or styles, and self-training. Methods like adversarial training \cite{chen2018domain, hsu2020every} and style transfer \cite{chen2020harmonizing, inoue2018cross, kim2019diversify} aim to bridge domain gaps, but they struggle to balance feature discriminability. Self-training methods enhance performance on the target domain by utilizing source domain knowledge without relying on labels. For example, studies like \cite{deng2021unbiased}, \cite{li2022cross}, and \cite{cao2023contrastive} employ a mean-teacher framework, where a teacher network generates pseudo-labels to guide the student network in the target domain. Recently, \cite{do2024d3t} proposed a two-teacher, single-student architecture with zig-zag learning to reduce domain confusion. However, frequent teacher switching causes loss values to diverge, leading to model collapse on complex datasets. In contrast, our proposed SAGA augmentation, integrated with different adaptation algorithms, consistently enhances performance across multiple datasets. \\
\textbf{Multi-modal Object Detection dataset}
Datasets like VEDAI \cite{razakarivony2016vehicle} introduced RGB-IR imagery for object detection, featuring 12,000 bird’s-eye-view drone images across nine categories. The KAIST dataset \cite{hwang2015multispectral}, released the same year, focused on pedestrian detection for autonomous vehicles and gained significant traction despite challenges like misalignment caused by mechanical vibrations and calibration errors. While KAIST employed beam-splitter co-registration for alignment, other datasets like CVC14 \cite{gonzalez2016pedestrian} faced persistent alignment issues due to hardware limitations. In subsequent years, more extensive datasets have emerged, including FLIR \cite{fliresh}, LLVIP \cite{jia2021llvip}, DroneVehicle \cite{sun2020drone}, and M3FD \cite{liu2022target}. Drone-based detection presents distinct challenges, such as object occlusion and scale variations, which datasets like VEDAI and DroneVehicle aim to address.
The LLVIP dataset focuses on low-light surveillance for pedestrian detection, while M3FD facilitates multimodal fusion for detection using high-resolution imagery. However, LLVIP, DroneVehicle, and M3FD depend on manual RGB-IR image co-registration, a labor-intensive process unsuitable for real-time drone operations\cite{song2024misaligned}. Hardware-based alignment methods, such as calibration devices or sensors, often struggle due to environmental factors like temperature fluctuations, mechanical instability, and limitations in feature-matching algorithms\cite{brenner2023rgb}. 
Additionally, these datasets contain fewer samples, limited classes, multi-scale variations, and diverse slant angles.
To tackle these challenges, we present the IndraEye dataset, specifically designed to accommodate real-world constraints like long-tail distributions, occlusion, and scale diversity. A qualitative comparison with existing datasets is presented in Table \ref{Qual_comp}. \\

%% file: sec/4_methodology.tex
\section{Semantic Aware Gray color Augmentation (SAGA) for Domain Adaptation}
% The proposed Semantic Aware Gray color Augmentation (SAGA) and IndraEye dataset are introduced in the following sections. 
% Section 3.1 describes the Mean Teacher framework for unsupervised domain adaptation, adapted for domain-adaptive thermal object detection. Section 3.2 presents the SAGA augmentation method, followed by Section 3.3, which introduces the IndraEye dataset.
The following sections outline SAGA and IndraEye. Section 3.1 outlines the Mean Teacher framework for thermal object detection, Section 3.2 presents SAGA, and Section 3.3 introduces IndraEye.

\subsection{Preliminaries on the Mean Teacher Framework for Domain Adaptation }
% \subsection{Object Detection}
The Mean Teacher (MT) framework \cite{tarvainen2017mean} for domain-adaptive object detection leverages a teacher-student mutual learning framework to transfer knowledge from a labeled source domain to an unlabeled target domain \cite{deng2021unbiased, deng2023harmonious, liu2021unbiased}. Specifically, in the context of domain-adaptive thermal object detection, the source domain comprises RGB images, while the target domain consists of infrared (IR) images.
The teacher and student models are two object detection networks with identical architectures. This framework processes source and target data simultaneously, enabling mutual knowledge transfer. The teacher model, trained on the labeled source domain, generates pseudo labels for the unlabeled target domain data. The student model learns from these pseudo labels, and its updated weights are periodically transferred to the teacher model, enabling continuous improvement. The framework is optimized using the following loss function:
\begin{equation}
    \textit{L} = \textit{L}_{src} + \textit{L}_{tgt}
\end{equation}
where, $L_{src}$ denotes the loss associated with the source domain, while $L_{trt}$  represents the loss for the target domain. The teacher model's weights are updated using the student model's weights through the Exponential Moving Average (EMA) mechanism. Throughout the training process, the teacher model update effectively forms an ensemble of the student model's weights, represented as:
\begin{equation}
    \theta^{T} =  \alpha \theta^{T} + (1-\alpha)\theta^{S}
\end{equation}
where, $ \theta^{T}$   represents the teacher model's weights, while 
$ \theta^{S}$   denotes the student model's weights. CMT \cite{cao2023contrastive} introduces a framework that integrates contrastive learning with the MT methodology, where the student network is trained using stronger augmentations, while the teacher model is trained with weaker augmentations. D3T \cite{do2024d3t}, on the other hand, employs two separate teacher models followed by a zigzag training mechanism. For additional information, please refer to \cite{cao2023contrastive,do2024d3t}.

\subsection{Semantic Aware Gray color Augmentation (SAGA) for Domain Adaptation}
 
Color and structure serve as fundamental elements of an image, both essential for object detection models, with the impact of color bias extensively studied through an emphasis on structural features \cite{hou2020learning, gong2024exploring}. Models trained on RGB data often heavily depend on these color cues, as chromatic diversity is crucial for distinguishing objects in RGB imagery \cite{zhang2022fmcnet}. However, infrared (IR) imaging modalities inherently lack the ability to capture rich color and structural details compared to RGB, posing significant challenges for cross-modal analysis and domain adaptation \cite{zhao2021joint, cheng2023multi}. In the target IR domain, the absence of chromatic diversity leads to a significant reduction in color information, which can contribute to an increased rate of false positives. Domain-adaptive thermal object detection strives to balance the dependence on RGB color bias while retaining essential structural features, in the target IR domain \cite{wu2017rgb}. To mitigate this bias while preserving essential visual cues, we introduce Semantic-Aware Gray Color Augmentation (SAGA), a simple, effective augmentation easily integrated into any object detection algorithm.

When training an object detection model on the source domain using RGB images, let the dataset be represented as \textit{S}=\{$x_i$, $y_i$\}, where $x_{i}$ denotes the images and $y_{i}$  represents the corresponding object labels. These labels are utilized to selectively convert object instances in $x_{i}$ to grayscale while preserving the original color information in the background, rather than applying grayscale conversion to the entire image. SAGA selectively converts object instances to grayscale while retaining the background's color information, ensuring semantic awareness.

\begin{figure}[]
    \centering
    \includegraphics[scale=0.415]{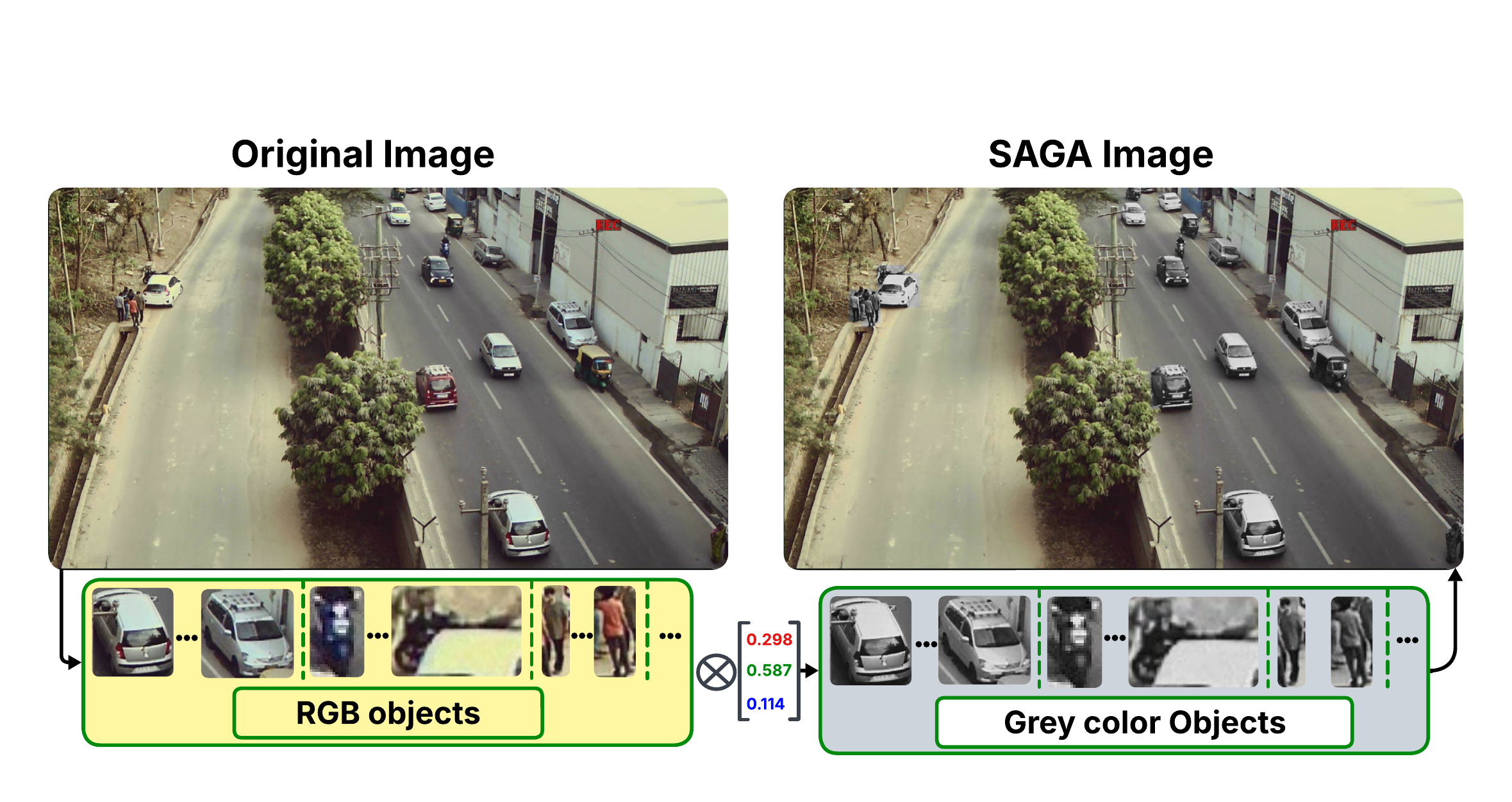}
    \setlength{\abovecaptionskip}{-11pt} 
    \setlength{\belowcaptionskip}{-15pt}
    \caption{Illustration of SAGA augmentation. The process involves extracting objects from the image, converting them to grayscale, and reintegrating them into the original image while preserving background color information.}
    \label{fig:sagav1}
\end{figure}

\begin{figure}[]
    \centering
    \includegraphics[scale=0.155]{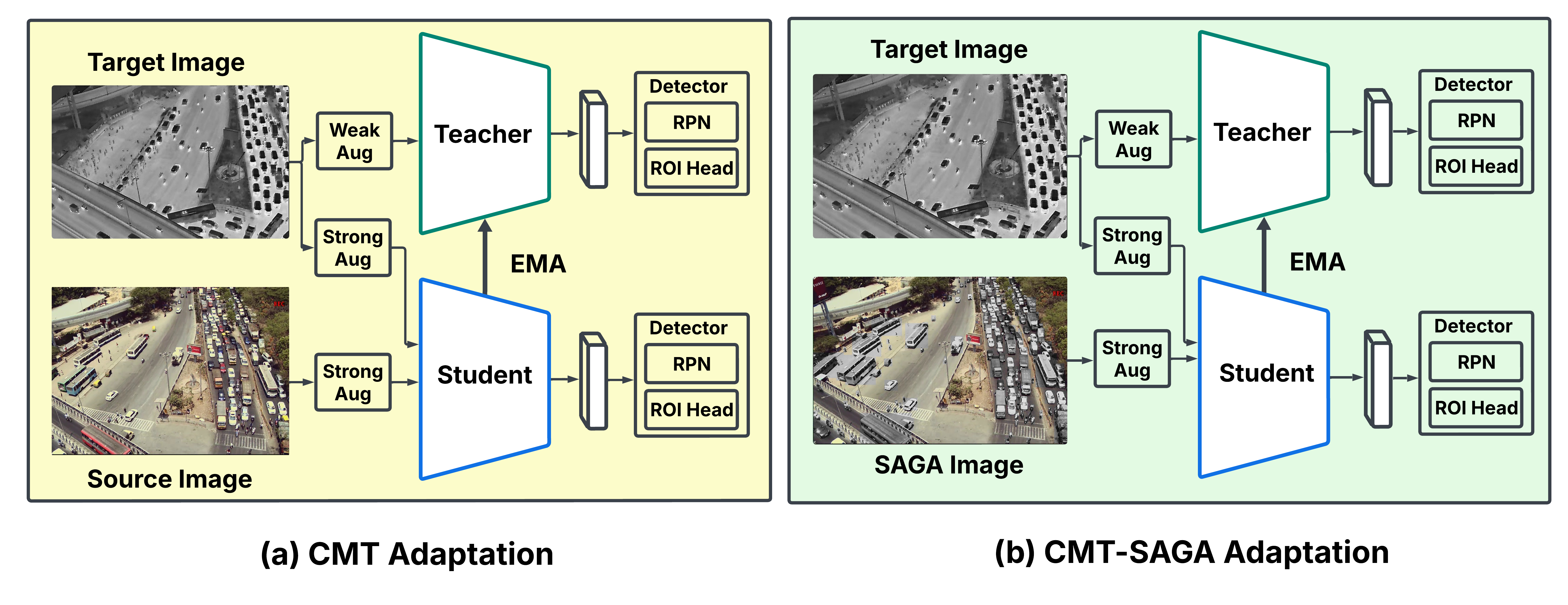}
    \setlength{\abovecaptionskip}{-5pt} 
    \setlength{\belowcaptionskip}{-11pt}
    \caption{Domain-adaptive thermal object detection with RGB as the source domain and IR as the target domain. (a) Vanilla CMT on the IndraEye dataset. (b) CMT with SAGA on the IndraEye dataset.}
    \label{fig:cmt_saga}
\end{figure}

Consider an image \textit{I} containing \textit{n} objects, represented as \textit{I} = \{$o_1$, $o_2$, ..., $o_n$\}, where $o_1$, $o_2$, and $o_n$ denote the individual objects within the image. Each object is extracted, converted to grayscale, and then reintegrated into the original image, ensuring that only the object instances appear in grayscale while the background retains its original colors. This process produces a hybrid RGB-grayscale image, as illustrated in Fig. \ref{fig:sagav1}, while preserving the structural integrity of the objects. SAGA can be represented as the following equation, which converts given image \textit{I} by multiplying the red, green and blue channel of each pixels by the values [0.2989, 0.587, 0.114] respectively,
\begin{equation}
    I =  I \times [0.2989, 0.5870, 0.1140]
\end{equation}
RGB-to-IR adaptation highlights that directly transferring knowledge from the RGB source domain to the IR target domain can lead to the transfer of irrelevant features via the EMA technique \cite{do2024d3t}, as there is no explicit mechanism to selectively extract meaningful information. This challenge contributes to an increase in false positives within the pseudo labels generated by the teacher network. To mitigate this, SAGA is integrated into the mean-teacher framework, enabling effective knowledge transfer from labeled RGB images to the unlabeled IR target domain. SAGA is used for domain-adaptive thermal object detection, where source images undergo instance-level gray augmentation to enhance feature adaptation before being trained within their respective frameworks, as shown in Fig. \ref{fig:cmt_saga}.

\begin{table}[ht]
\centering

\begin{tabular}{c c c l}
\toprule
\multicolumn{1}{l}{\textbf{Sensor}} & \multicolumn{1}{l}{\textbf{Resolution}} & \multicolumn{1}{l}{\textbf{Wavelength}} & \textbf{FoV} \\ 
\midrule
RGB camera                        & 1280x720                                 & 400-700\text{nm}                                & 60$^\circ$                      \\ 
IR Camera                        & 640x480                                  & 8-14$\mu$m                                     &  32$^\circ$                      \\ 
\bottomrule
\end{tabular}
\caption{Sensor specifications of the DragonEye 2 RGB-IR camera used for the IndraEye dataset.}
\label{sensor_specs}
\end{table}

% To create the IndraEye dataset, we used the DJI M600 Pro drone (Fig. 1), equipped with a DragonEye2 camera mounted on a gimbal to capture various slant-angle views by adjusting pitch and yaw. For low-altitude captures, the camera and gimbal setup were placed on a 3-meter elevated tripod. The DragonEye2 camera includes one Electro-Optical (EO) sensor and one uncooled Infrared (IR) sensor, with detailed specifications provided in Table 2. IndraEye was collected at different times of day—noon, evening, and night—to account for variations in illumination and backgrounds, thereby reducing inherent biases in neural networks (additional examples are available in the supplementary material). After image collection, the dataset was constructed by manually annotating target objects with bounding boxes in both EO and IR data.

% Videos from the EO-IR camera were recorded at seven locations in Bengaluru, including the Indian Institute of Science campus. 
% Dataset at each location was collected for approximately 4 minutes, resulting in about 20,000 frames per scene. To improve object diversity and manage redundant content, every 35th frame was selected for inclusion in the dataset. 
\subsection{Aerial Multi-Modal Perception Dataset}
\textbf{Acquisition process} The IndraEye dataset was collected using a DJI M600 Pro drone equipped with a multi-sensor RGB-IR camera mounted on a gimbal, allowing for image capture from various slant angles by dynamically adjusting the pitch and yaw. The dataset also consists of data collected by mounting the RGB-IR camera on a 3 meter elevated tripod placed over places such as bridge and underpass, there by accounting for low range slant angle that is similar to that of a traffic pole mounted camera images. The information about the sensor can be found in Table \ref{sensor_specs}. IndraEye is one of the very few mutlispatial and multitemporal slant angle dataset available at two modalities i.e. RGB and IR. The dataset was manually structured by annotating the potential objects of interest with bounding boxes in RGB and IR images. The video was shot for approximately 4 to 5 minutes in each of the different locations as shown in Table \ref{table: dataset description_class}. This generated around 20,000 frames per scene as the refresh rate of the RGB camera is 50Hz and that of IR is 10Hz. To avoid redundancy of objects we consider only 35th frame in the video to obtain diverse object rich frames. \\
\textbf{Camera calibration} Recent work \cite{brenner2023rgb} has shown that calibrating RGB and IR camera on a drone vehicle under motion is challenging and not feasible due to the variations in camera parameters such as focal length, field of view (FOV) and refresh rate during movement of the drone. This is a serious issue since most of the camera calibration methods requires computing homography matrix and running feature matching algorithm such as SIFT\cite{lowe2004distinctive}, SURF\cite{bay2006surf} on the device to align both RGB and IR images. This process limits edge-based machine learning due to computational constraints \cite{brenner2023rgb}. Stereo camera calibration is effective but challenged by differing RGB-IR FOVs, leading to depth-dependent parallax \cite{gao2021stereo}. This occurs due to variations in sensor parameters and viewing angles, causing objects at different depths to appear in different image positions. \\
% Due to the hardware and software limitations of the camera, calibrating a system like DragonEye2 where both EO and IR cameras are integrated into a single module mounted on a drone has proven to be challenging \cite{ADD_CITATION}. The difficulty is mainly due to the variations in the camera parameters such as focal length and field of view during the movement of the drone vehicle.
% (because of the variations caused by the focal length of the sensor for better clarity on the object features) the field of view (FoV) varies dynamically with the movement of the drone along with the camera,
% This causes complications in the calibration of the EO and IR sensors as reported by \cite{THE_REVIEW_PAPER}.
% complicating the calibration of both cameras as mentioned in \cite{brenner2023rgb}. 
%While stereo camera calibration is effective, it encounters a significant challenge due to different FOV's of the modalities (RGB-IR) which can result in a parallax effect that varies at different depths \cite{gao2021stereo}. This phenomenon is because of the difference variation in sensor parameters and viewing angles between the cameras, causing objects at various depths to appear at different positions in the image. 
% Consequently, employing a single homography matrix, a transformation that maps points from one image to corresponding points in another is not very effective due to the variation in perspective. 
This misalignment becomes evident in the fused data when the vehicle with the camera is in motion \cite{brenner2023rgb}. Image alignment can lead to inaccuracies, particularly with long distance and small objects. As a result, the process of co-registration is avoided intentionally. While the RGB and IR images are captured with the same timestamp, they still exhibit minor misalignment due to change in refresh rate of the sensor. This limitation highlights the importance of multimodal approaches that eliminate the necessity for co-registered images, such as domain adaptation \cite{do2024d3t}, \cite{gan2023unsupervised}.\\
\textbf{Statistics of the dataset} 
IndraEye comprises 5,612 images that includes multiple viewing angles, altitudes, backgrounds, and times of day. The RGB images are divided into 2,336 samples (2,026 for training, 60 for validation, and 250 for testing), while the IR images are divided into 3,276 samples (2,973 for training, 58 for validation, and 245 for testing). The dataset additionally includes day and night splits. The goal was to create a diverse and comprehensive dataset with both RGB and IR images, suitable for various conditions and contexts.
IndraEye features 13 classes: backhoe loader, bicycle, bus, car, cargo truck, cargo trike (a medium-sized three-wheeled cargo vehicle), ignore, motorcycle, person, rickshaw (a small three-wheeled passenger vehicle), small truck, truck, tractor, and van. More information regarding the dataset is provided in 
% Detailed class distribution for the IndraEye training set is shown in
Table \ref{table: dataset description_class}. To improve object diversity and reduce redundancy, every 35th frame was extracted from the 20,000 frames captured per scene.\\

\textbf{Annotation}  For the IndraEye dataset, we provide ground-truth labels for both object detection and pixel-level semantic segmentation across all classes using a two-stage approach. 
\begin{enumerate}
  \item  \textbf{Zero-Shot Annotations with Human in the Loop:}
    In this initial step, we generate zero-shot annotations by utilizing pre-trained models. These models are used to produce preliminary annotations with human oversight, ensuring that the annotations align closely with the true object locations and characteristics.
    \item \textbf{Manual Verification and Refinement:}
    After generating the initial annotations, each label undergoes thorough manual verification. This step involves a detailed review of the annotations to correct any discrepancies and refine the labels for higher accuracy.
\end{enumerate}

This structured approach ensures that both object detection and pixel-level annotations are meticulously crafted and reliable. For the object detection task, RGB and IR images are manually annotated with bounding boxes using the X-AnyLabeling tool \cite{X-AnyLabeling}. This tool facilitates efficient annotation by allowing users to load pre-trained models, such as those from the YOLO family of networks. After multiple iterations of correction and evaluation, highly precise annotations have been generated.  For semantic segmentation, we use SAMRS \cite{wang2024samrs}, an extension of SAM \cite{kirillov2023segment}, optimized for aerial imagery. SAMRS leverages remote sensing datasets to efficiently generate masks for large-scale segmentation tasks. 
This approach has streamlined our process for producing pixel-level annotations for IndraEye. The annotations for both tasks are carefully designed to match the established class schema.\\
\textbf{Ethics and policy} The proposed dataset undergoes a thorough manual review to ensure privacy protection. Faces of individuals and vehicle license plates that are clearly visible in the images are blurred to maintain confidentiality and safeguard personal rights. This process ensures that individuals' privacy is upheld while preserving the dataset's utility.

%% file: sec/5_experimental.tex
\section{Experiments}
In this section, we present the experimental results of the proposed SAGA on state-of-the-art unsupervised domain adaptation models, including D3T \cite{do2024d3t}, CMT \cite{cao2023contrastive}, and AT \cite{li2022cross}. These models are evaluated on two open-source datasets, FLIR \cite{zhang2020multispectral} and LLVIP \cite{jia2021llvip}, along with the proposed IndraEye dataset. We also assess object detection models across various settings to establish a benchmark on the IndraEye dataset.

% \textbf{How is this??}
% In this section, we demonstrate the effectiveness of our proposed augmentation SAGA for Unsupervised RGB-T domain adaptation by comparing with recent state-of-the-art algorithms and datasets. We also demonstrate the importance of aerial RGB-T dataset for object detection mainly for approaches like domain adaptation. We show that SAGA with the state-of-the-art unsupervised domain adaptation technique beats the networks without SAGA by a significant margin as shown in the results.

% \begin{figure*}[t]
%     \centering
%     \includegraphics[scale=0.3]{images/pred.png}
%     \caption{Output predictions to higlight the importance of the SAGA augmentation.}
%     \label{fig:pred}
% \end{figure*}

% Decrease the size of this font
% \subsection{Datasets}
% % \textbf{Dataset and Evaluation Metrics}
% 

% \textbf{How is this??}

\subsection{Datasets and Evaluation metric }
%\textbf{Datasets: } We perform all the experiments for SAGA on three RGB-T datasets two of which are publicly available and are used for object detection task i.e. FLIR-ADAS and LLVIP and thirdly we evaluate all the methods on our dataset to set the benchmark. FLIR-ADAS dataset has 9,214 thermal and RGB image pairs and has mainly three class labels car, person, and bicycle. The dataset contains images of resolution 640 x 512 and are collected during both day and night images. All experiments are carried out on the official training and testing split provided. To demonstrate the effect of lighting condition, LLVIP dataset is chosen and consists of only night-time images. It has 15,488 RGB-T image pairs and each image has a resolution of 1080x720 pixels and LLVIP has only person class. IndraEye dataset consists of 5,612 RGB and 3276 thermal images collected during all times of the day and night. And notably it has 13 vehicle objects i.e. car, backhoe loader, bus, van, bicycle, cargo truck, ignore, cargo trike, small truck, motorcycle, person, tractor, truck and rickshaw. \\
\textbf{Datasets}  Experimental evaluation is conducted using the FLIR, LLVIP, and IndraEye datasets. FLIR \cite{zhang2020multispectral} is collected from an autonomous driving perspective. This dataset consists of 5,142 aligned RGB-IR image pairs, with 4,129 images allocated for training and 1,013 for testing, covering daytime and nighttime scenes. LLVIP \cite{jia2021llvip} is a low-light vision dataset captured from a slant angle, roughly three stories above ground. It contains 15,488 RGB-IR image pairs, mainly featuring extremely dark scenes and focusing solely on the Pedestrian class.
IndraEye is a slant-angle dataset with 2,336 RGB and 3,276 IR images. Captured from diverse backgrounds and angles, it covers 13 classes, including vehicles and pedestrians, with day and night scenes under varying illumination.\\
\textbf{Evaluation metrics} For experimental validation, we evaluate all the networks using the Mean Average Precision (mAP) score, calculated from precision and recall scores. An mAP threshold of 0.5 is used for all experiments. \\
% \begin{table}[]
% \centering
% \fontsize{9pt}{9pt}\selectfont
% \begin{tabular}{|c|c c c|}
% % \toprule
% \hline
% Algorithm\textbackslash{}Dataset & FLIR & LLVIP & IndraEye \\
% % \midrule
% \hline
% D3T                              & 61.8 & 88.1  & 30.3     \\
% % \midrule
% \hline
% D3T*                             & \textbf{66.1} & \textbf{88.5}  & \textbf{32.9}     \\
% % \midrule
% \hline
% CMT                              & 63.3 & 41.6  & 17.1     \\
% % \midrule
% \hline
% CMT*                             & \textbf{65.7} & \textbf{49.2}  & \textbf{19.3}     \\
% % \midrule
% \hline
% AT                               & TR     & TR      &TR          \\
% % \midrule
% \hline
% AT*                              &  TR    &  TR     &  TR        \\
% % \midrule
% \hline
% Source                           &  37.8    &  82.8     & \textbf{50.5}         \\
% % \midrule
% \hline
% Oracle                           & 67.5     &  83     &   \textbf{65.5}       \\
% % \bottomrule
% \hline
% \end{tabular}
% \caption{Performance comparison with state-of-the-art methods on FLIR, KAIST and our proposed dataset \newline * indicates with the propose augmentation strategy on the respective algorithms}
% \label{table:UDA_results}
% \end{table}
%\vspace{0.5em}
%\hspace{-1em}{}
\textbf{Implementation Details} 
For evaluating domain adaptation networks, we train the models using strategies from \cite{do2024d3t}, \cite{cao2023contrastive}, and \cite{li2022cross}. During the burn-up stage, the student network trains on the source domain for 10,000 iterations. Adaptation to the target domain occurs in subsequent iterations, with a total of 50,000 iterations on the FLIR and LLVIP datasets. For the IndraEye dataset, we train all algorithms for 20,000 iterations, as the D3T \cite{do2024d3t} network diverges after this point, leading to poor results. We hypothesize this is due to the dataset's complexity and the frequent alternation between the student and teacher networks during training (zig-zag training). All experiments in \ref{table:UDA_results} used a batch size of 16 on four NVIDIA V100 GPUs with a VGG16 backbone and the same hyperparameters as the respective works. \\
Object detection models, including FasterRCNN, ReDet, and ORCNN, were benchmarked on the IndraEye dataset using four Nvidia V100 GPUs with a batch size of 6.
We employed the SGD optimizer with a learning rate of 0.0025, weight decay of 0.0001, and momentum of 0.9, training for 12 epochs. The YOLOv8\cite{yolov8_ultralytics} model was trained with a learning rate of 0.00001, a batch size of 16, and an image size of 640 for a total of 100 epochs.

\begin{table}[]
\centering
\resizebox{3.2in}{!}{
\begin{tabular}{@{}c c c c c c@{}}
\toprule
% \rowcolor[HTML]{9B9B9B} 
\textbf{Models} & \textbf{Train on RGB} & \textbf{Test on RGB} & \textbf{Train on IR} & \textbf{Test on IR} & \textbf{mAP50} \\ 
\midrule
FasterRCNN \cite{NIPS2015_14bfa6bb}     & \checkmark                   & \checkmark                  & \ding{55}                   & \ding{55}                  & 47.6 \\ 

ReDet \cite{han2021redet}          & \checkmark                   & \checkmark                  & \ding{55}                   & \ding{55}                  & 43.9 \\ 

ORCNN \cite{xie2021oriented}          & \checkmark                   & \checkmark                  & \ding{55}                   & \ding{55}                  & 56.3 \\ 

Faster RCNN \cite{NIPS2015_14bfa6bb}    & \ding{55}                   & \ding{55}                  & \checkmark                   & \checkmark                  & 57.3 \\ 

ReDet \cite{han2021redet}          & \ding{55}                   & \ding{55}                  & \checkmark                   & \checkmark                  & 43.8 \\ 

ORCNN \cite{xie2021oriented}          & \ding{55}                   & \ding{55}                  & \checkmark                   & \checkmark                  & 65.6 \\ 
\bottomrule
\end{tabular}}
% \label{fig:detection-dota11}
% \caption{Add citations and table legend.}
% \end{table}
\caption{Performance of object detection algorithms on the proposed IndraEye dataset.}
\label{table:obj_det benchmark ie}
\end{table}

\subsection{Experimental Results}

% \centering
% \begin{figure*}[h]
%     \centering
%     \includegraphics[scale=0.31]{images/cmt_pred.png}
%     \caption{Output predictions to highlight the importance of the SAGA augmentation. (a) RGB image and (b) IR image from the dataset. (c) Demonstration of the challenges faced by a state-of-the-art algorithm on the dataset. (d) Significant reduction in false positives (highlighted with red circles in (c) and (d)) achieved through the proposed SAGA augmentation, showcasing its effectiveness}
%     \label{fig:pred}
% \end{figure*}

\subsubsection{Unsupervised Domain Adaptation}
As observed in table \ref{table:UDA_results}, the proposed SAGA consistently enhances the performance of both CMT and D3T across all three datasets. Specifically, it improves CMT by 2.4 mAP on FLIR, 3.6 mAP on LLVIP, and 2.2 mAP on IndraEye respectively. Similarly, it enhances the performance of D3T by 4.29 mAP on FLIR. Similar trends are observed in other settings. This clearly demonstrates that transferring all information from the student to the teacher is not beneficial. Instead, mitigating color bias while preserving semantic information leads to consistently better performance. This makes the proposed augmentation technique highly effective for RGB-IR domain adaptation, as it consistently improves performance across different look angles, which is crucial for drone imagery. \\
\textbf{Compatibility with existing techniques.} Table \ref{table:abalation} demonstrates that SAGA is compatible with various existing techniques along with domain adaptation approaches. For methods designed to enhance the representation capacity of encoders, such as Jigen\cite{carlucci2019domain}, SAGA does not negatively impact training despite being an augmentation technique. Instead, it complements these methods, supporting their learning and improving the representation ability of models for RGB-IR adaptation. \\
\textbf{Comparison with augmentation techniques:}Although grayscale augmentation helps align RGB to Thermal for domain adaptation, we show that full grayscale augmentation reduces CMT model performance on the FLIR dataset, as observed in table \ref{table:abalation_grey}.
Converting the entire image to grayscale removes important local details, like fine boundaries, making it difficult for the model to identify key features. In contrast, instance-level grayscale augmentation retains these details, enhancing performance.\\
\begin{figure}[htbp]
    \centering
    \includegraphics[scale=0.35]{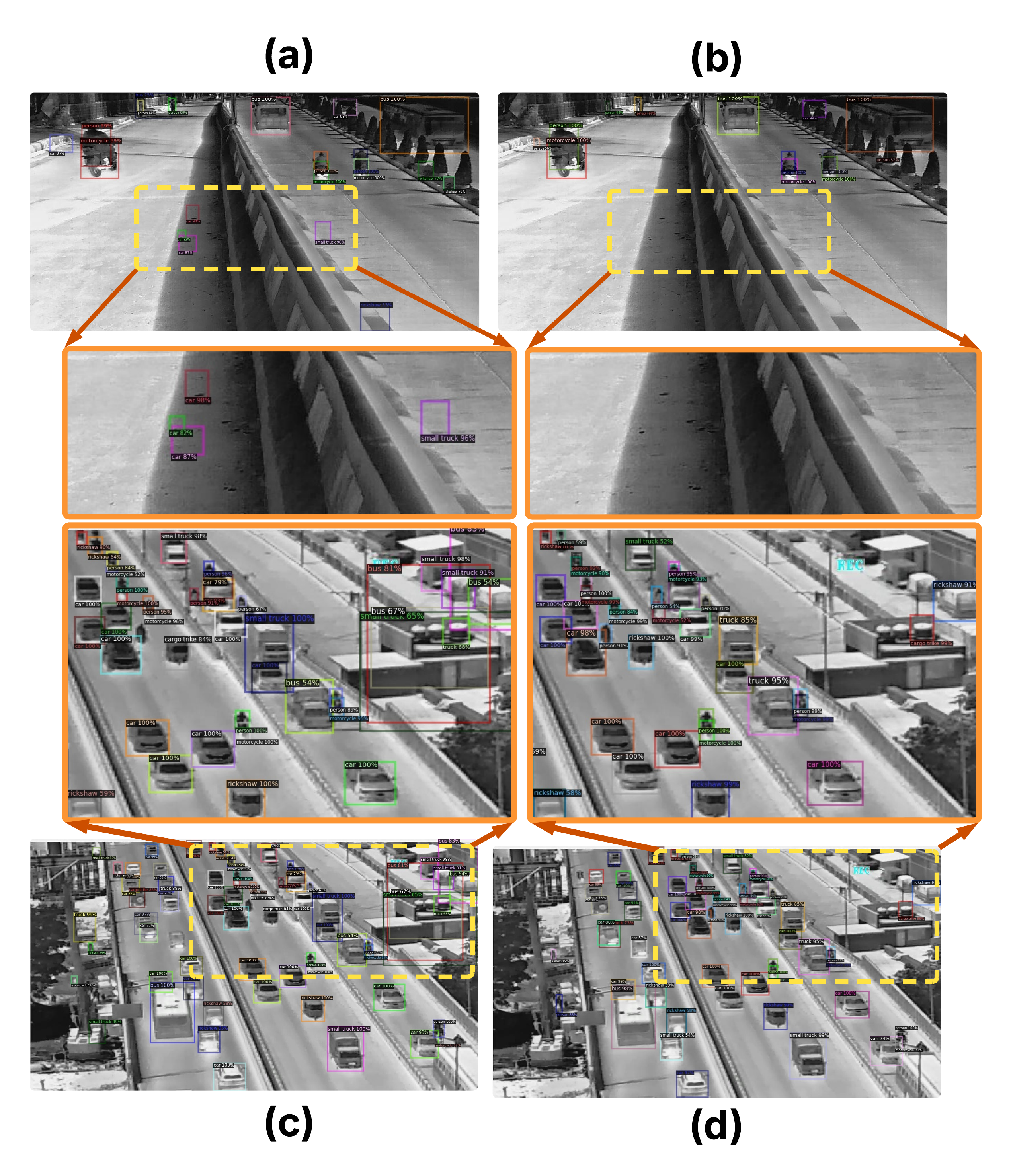}
    \setlength{\abovecaptionskip}{-1pt}
    \setlength{\belowcaptionskip}{-8pt}
    \caption{Output predictions to highlight the importance of the SAGA augmentation on CMT algorithm. (a) and (c) shows the increase in false positives while using vanilla CMT. Meanwhile (b) and (d) shows the reduction in false positives when using SAGA with CMT, showcasing its effectiveness.}
    \label{fig:pred_1}
\end{figure}
\textbf{Qualitative results}  Qualitative predictions show improved performance for small-scale objects (Fig. \ref{fig:pred_1}), a common challenge in IR imagery. Unlike RGB, IR sensors lose information from certain viewpoints, especially when objects are viewed from different angles. Grayscale augmentation helps address this issue. We observe a significant reduction in false positives in the vanilla CMT model, and incorporating the proposed augmentation effectively mitigates this problem, while consistently improving performance across three benchmarked datasets, as shown in Table \ref{table:UDA_results}.

\begin{table}[h]
% \caption{Table illustrating how the features of the two modalities complement each other, utilizing a day-night split within the train-test split of the IndraEye dataset.}
% \label{table:algo}
\centering
% \fontsize{9pt}{9pt}
\begin{tabular}{l l l l}
% \rowcolor[HTML]{9B9B9B}
\toprule
\textbf{Algorithm} & \textbf{FLIR} &   \textbf{LLVIP} &  \textbf{IndraEye} \\
\midrule
% \vspace{-0.2em}
Source       & 37.8   & 82.8     & 50.5       \\ 
% \vspace{-0.2em}
\midrule
% \vspace{-0.2em}
Oracle       & 67.5   & 83     & 65.5       \\ 
% \vspace{-0.2em}
\midrule
D3T    & 61.8     & 88.1   & 30.3          \\ 
% \vspace{0.1em}
D3T + SAGA   & \textbf{66.1}   \textcolor{forestgreen}{$\uparrow4.3$}  &  \textbf{88.5} \textcolor{forestgreen}{$\uparrow0.4$}   & \textbf{32.9}  \textcolor{forestgreen}{$\uparrow2.6$}    \\ 
% \vspace{0.1em}
CMT    & 63.3   & 41.6     & 17.1      \\ 
% \vspace{0.1em}
CMT + SAGA   & \textbf{65.7}  \textcolor{forestgreen}{$\uparrow2.4$}   & \textbf{49.2}   \textcolor{forestgreen}{$\uparrow7.6$}    &  \textbf{19.3}  \textcolor{forestgreen}{$\uparrow2.2$}      \\ 
% \vspace{0.1em}
% AT       & 48.13   & TR     & 30.4       \\ 
% \vspace{0.1em}
% AT + proposed      & TR   & 30.0     & 30.4       \\ 
% \vspace{0.1em}
% \vspace{0.1em}

\bottomrule
\end{tabular}
% \caption{Domain adaptation ablation experiments - Jigen and all settings, on .. dataset. - check table and figure legends, is it above or below.}
\caption{Experimental Results on State-of-the-art algorithms on two standard datasets i.e, FLIR and LLVIP. Source refers to the accuracy obtained by training on RGB domain and testing on IR domain, while Oracle refers to training and testing on the same i.e. IR domain.}
\label{table:UDA_results}
\end{table}
\subsubsection{Benchmarking on IndraEye dataset}
{\textbf{Object Detection}}
Detection models like ReDet \cite{han2021redet}, ORCNN \cite{xie2021oriented}, and Faster-RCNN \cite{NIPS2015_14bfa6bb} show limited in-domain performance (training and testing on the same modality), falling short of real-world, safety-critical application requirements, as shown in Table \ref{table:obj_det benchmark ie}. This can be attributed to factors like scale variability, varying illumination and background conditions, and challenges specific to RGB sensors, which are common in drone imagery. This highlights the significance of the IndraEye dataset and the need for model architectures that are generalizable and adaptable to diverse scenarios.  \\
Table \ref{table:detection-aerial} evaluates the performance of Yolov8x \cite{Jocher_Ultralytics_YOLO_2023} on IndraEye on five different configurations. The four configurations include: first, training on the RGB modality with daytime images and testing on nighttime images under low illumination conditions, highlighting the limitations of RGB imagery in low-light environments. The second configuration assesses the effectiveness of using only RGB sensors for aerial perception, where training incorporates both day and nighttime images, but inference is conducted solely under nighttime conditions. The third configuration is completely based on the capability of the model learning from the IR imagery. Precisely, in this setting IR modality with both day and nighttime images are used to train the DNN. As can be seen from the table \ref{table:detection-aerial}, comparison from the first three test configurations, it is clear that the IR modality is better suited for low-illumination conditions. The fourth and fifth configurations are counterparts to the second and third, evaluated under well-illuminated (daytime) conditions. Results show that RGB sensors, capturing rich texture, perform better in these conditions, while IR sensors excel in low-light environments due to their ability to capture thermal emissivity. \\
To benchmark other aerial perception datasets against IndraEye in the RGB domain, we train YOLOv8x on the VisDrone dataset, which has more training image instances than IndraEye. For a fair comparison, we focus on the common classes (5 classes) between both datasets. Despite VisDrone's larger training set, the cross-domain gap (between IndraEye to VisDrone and VisDrone to IndraEye) reveals that the IndraEye dataset helps the model generalize better, as shown in Table \ref{table:visdrone_table}. We attribute this improvement to the greater diversity and higher quality annotations in the IndraEye dataset.

\begin{table}[]
\label{EO-D+N}
\centering
\begin{tabular}{l l l}
% \rowcolor[HTML]{9B9B9B}
\toprule
\textbf{\begin{tabular}[c]{@{}c@{}}Train Setting\end{tabular}} & \textbf{\begin{tabular}[c]{@{}c@{}}Test Configuration\end{tabular}} & \textbf{mAP50} \\ 
\midrule
RGB-Day   & RGB-Night    & 30.0   \\ 
RGB-Day + RGB-Night  & RGB-Night  & 52.0  \\ 
IR-Day + IR-Night  & IR-Night  & 73.6   \\ 
RGB-Day  + RGB-Night  & RGB-Day   & 90.7    \\ 
IR-Day + IR-Night & IR-Day    & 77.0   \\ 
\bottomrule
\end{tabular}
\caption{Table illustrating the features of the two modalities complementing each other, utilizing a day-night split within the train-test split of the IndraEye dataset.}
\label{table:detection-aerial}
\end{table}

\hspace{-1em}

\begin{table}[h]
\centering
\fontsize{9pt}{9pt}
\label{DG}
% \fontsize{7pt}{7pt}\selectfont
% \scriptsize   # This is to resize the table to smaller dimension
% \resizebox{1.7in}{!}{
% \begin{tabular}{@{}lll@{}}
% \toprule
% \textbf{Train set} & \textbf{Test set} & \textbf{mAP50} \\ 
% \midrule
% VisDrone           & VisDrone \textbf{cite}         & 53.3           \\ 
% IndraEye           & VisDrone          & 32.5           \\ 
% IndraEye           & IndraEye          & 83.135         \\ 
% VisDrone           & IndraEye          & 52.085         \\
% \bottomrule
% \end{tabular}}
\begin{tabular}{lll}
\toprule
\textbf{Train set} & \textbf{Test set} & \textbf{mAP50} \\ 
\midrule
VisDrone           & VisDrone \cite{zhu2021detection}        & 53.3           \\ 
IndraEye           & VisDrone          & 32.5  \hspace{0.5em}  \textcolor{red}{$\downarrow20.8$}        \\ 
IndraEye           & IndraEye          & 83.1         \\ 
VisDrone           & IndraEye          & 52.1    \hspace{0.5em}  \textcolor{red}{$\downarrow31.05$}    \\
\bottomrule
\end{tabular}
\caption{Generalization of object detection model trained on Visdrone RGB imagery datasets.}
\label{table:visdrone_table}
\end{table}

%% file: sec/6_abalation.tex
%\section{Ablation}

% \begin{table}[h]
% \centering
% \begin{tabular}{@{}l l l l l l@{}}
% \toprule
% \textbf{Model}  &  \textbf{Vanilla} & \textbf{Full gray} & \textbf{SAGA} &  \textbf{Jigen} &  \textbf{mAP} \\ \midrule
% \multirow{3}{*}{CMT}  &  \checkmark     &       &      &     & 63.3 \\
%       &    \ding{55}    &   \checkmark      &     &     & 61.1 \\
%       & \ding{55}     &   \ding{55}  &  \checkmark     &     & 65.7 \\ 
%       & \ding{55}   & \ding{55}   & \ding{55}  & \checkmark   & 68.2 \\ \bottomrule
% \end{tabular}
% \caption{Table illustrating the ablation study of SAGA augmentation and Jigen Pretraining effect on CMT architecture}
% \label{table:abalation}
% \end{table}

\begin{table}[h]
\centering
\begin{tabular}{@{}l l l l l l@{}}
\toprule
\textbf{Model}  &  \textbf{Vanilla} & \textbf{Full gray} & \textbf{SAGA} &  \textbf{mAP50} \\ \midrule
\multirow{3}{*}{CMT}  &  \checkmark     &       &     & 63.3 \\
      &    \ding{55}    &   \checkmark      &     & 61.1 \\
      & \ding{55}     &   \ding{55}  &  \checkmark    & 65.7
      % & \ding{55}   & \ding{55}   & \ding{55}  & \checkmark   & 68.2 
      \\ \bottomrule
\end{tabular}
\caption{Table illustrating the ablation study of SAGA augmentation and Jigen Pretraining effect on CMT architecture.}
\label{table:abalation_grey}
\end{table}

\begin{table}[h]
\centering
\begin{tabular}{@{}l l l@{}}
\toprule
\textbf{Model}  &   \textbf{mAP50} \\ \midrule
Oracle      & 63.3 \\
Source  & 61.1 \\  \midrule
D3T  & 61.8 \\
D3T+SAGA  &  66.1 \\
D3T + Jigen  &  61.5  \\  \midrule
D3T + SAGA + Jigen  &  \textbf{68.2} \hspace{0.3em} \textcolor{forestgreen}{$\uparrow6.4$} \\  \midrule
CMT  & 65.7 \\ 
CMT + SAGA  &  65.7  \\
CMT + Jigen  &  64.7  \\ \midrule
CMT + SAGA + Jigen  &  \textbf{68.3} \hspace{0.3em}  \textcolor{forestgreen}{$\uparrow2.6$} \\
\bottomrule
\end{tabular}
\caption{SAGA on state-of-the-art domain adaptation methods, highlighting its compatibility with existing techniques like Jigen.}
\label{table:abalation}
\end{table}

%% file: sec/7_conclusion.tex
\section{Conclusion}
In this paper, we present Semantic-Aware Gray color Augmentation (SAGA), a novel technique that mitigates color bias and bridges the domain gap by extracting relevant object-level features for IR images. By integrating SAGA with the mean-teacher framework, we enable efficient knowledge transfer from labeled RGB images to unlabeled IR domains thereby achieving consistent performance gains of +0.4$\%$ to +7.6$\%$ when integrated with SOTA domain adaptation methods. To tackle the scarcity of RGB-IR drone imagery datasets, we present IndraEye, the first multisensor aerial perception dataset tailored for diverse classes and illumination conditions. The dataset includes diverse background environments and scene characteristics, featuring 2,336 RGB images and 3,276 IR images across different viewing angles, altitudes, and times of day. Our experimental results demonstrate that SAGA significantly enhances RGB-to-IR adaptation for autonomous driving and the IndraEye dataset. 

\section{Acknowledgments}
This work was supported by The Bengaluru Traffic Police (BTP), The Ministry of Electronics and Information Technology (MeitY) \& The Kotak IISc AI–ML Centre (KIAC).

%% file: main.bbl
\begin{thebibliography}{69}
\providecommand{\natexlab}[1]{#1}
\providecommand{\url}[1]{\texttt{#1}}
\expandafter\ifx\csname urlstyle\endcsname\relax
  \providecommand{\doi}[1]{doi: #1}\else
  \providecommand{\doi}{doi: \begingroup \urlstyle{rm}\Url}\fi

\bibitem[Bay et~al.(2006)Bay, Tuytelaars, and Van~Gool]{bay2006surf}
Herbert Bay, Tinne Tuytelaars, and Luc Van~Gool.
\newblock Surf: Speeded up robust features.
\newblock In \emph{Computer Vision--ECCV 2006: 9th European Conference on Computer Vision, Graz, Austria, May 7-13, 2006. Proceedings, Part I 9}, pages 404--417. Springer, 2006.

\bibitem[Brenner et~al.(2023)Brenner, Reyes, Susnjak, and Barczak]{brenner2023rgb}
Martin Brenner, Napoleon~H Reyes, Teo Susnjak, and Andre~LC Barczak.
\newblock Rgb-d and thermal sensor fusion: a systematic literature review.
\newblock \emph{IEEE Access}, 2023.

\bibitem[Cao et~al.(2023)Cao, Joshi, Gui, and Wang]{cao2023contrastive}
Shengcao Cao, Dhiraj Joshi, Liang-Yan Gui, and Yu-Xiong Wang.
\newblock Contrastive mean teacher for domain adaptive object detectors.
\newblock In \emph{Proceedings of the IEEE/CVF Conference on Computer Vision and Pattern Recognition (CVPR)}, pages 23839--23848, 2023.

\bibitem[Cao et~al.(2021)Cao, He, Wang, Wang, Yuan, Zhang, Zhang, Zhu, Van~Gool, Han, et~al.]{cao2021visdrone}
Yaru Cao, Zhijian He, Lujia Wang, Wenguan Wang, Yixuan Yuan, Dingwen Zhang, Jinglin Zhang, Pengfei Zhu, Luc Van~Gool, Junwei Han, et~al.
\newblock Visdrone-det2021: The vision meets drone object detection challenge results.
\newblock In \emph{Proceedings of the IEEE/CVF International conference on computer vision}, pages 2847--2854, 2021.

\bibitem[Carlucci et~al.(2019)Carlucci, D'Innocente, Bucci, Caputo, and Tommasi]{carlucci2019domain}
Fabio~M Carlucci, Antonio D'Innocente, Silvia Bucci, Barbara Caputo, and Tatiana Tommasi.
\newblock Domain generalization by solving jigsaw puzzles.
\newblock In \emph{Proceedings of the IEEE/CVF conference on computer vision and pattern recognition}, pages 2229--2238, 2019.

\bibitem[Chen et~al.(2020)Chen, Zheng, Ding, Huang, and Dou]{chen2020harmonizing}
Chaoqi Chen, Zebiao Zheng, Xinghao Ding, Yue Huang, and Qi Dou.
\newblock Harmonizing transferability and discriminability for adapting object detectors.
\newblock In \emph{Proceedings of the IEEE/CVF conference on computer vision and pattern recognition}, pages 8869--8878, 2020.

\bibitem[Chen et~al.(2018)Chen, Li, Sakaridis, Dai, and Van~Gool]{chen2018domain}
Yuhua Chen, Wen Li, Christos Sakaridis, Dengxin Dai, and Luc Van~Gool.
\newblock Domain adaptive faster r-cnn for object detection in the wild.
\newblock In \emph{Proceedings of the IEEE conference on computer vision and pattern recognition}, pages 3339--3348, 2018.

\bibitem[Cheng et~al.(2023)Cheng, Hua, Lu, Tu, Wang, and Wang]{cheng2023multi}
Ke Cheng, Xuecheng Hua, Hu Lu, Juanjuan Tu, Yuanquan Wang, and Shitong Wang.
\newblock Multi-scale semantic correlation mining for visible-infrared person re-identification.
\newblock \emph{arXiv preprint arXiv:2311.14395}, 2023.

\bibitem[Deng et~al.(2021)Deng, Li, Chen, and Duan]{deng2021unbiased}
Jinhong Deng, Wen Li, Yuhua Chen, and Lixin Duan.
\newblock Unbiased mean teacher for cross-domain object detection.
\newblock In \emph{Proceedings of the IEEE/CVF conference on computer vision and pattern recognition}, pages 4091--4101, 2021.

\bibitem[Deng et~al.(2023)Deng, Xu, Li, and Duan]{deng2023harmonious}
Jinhong Deng, Dongli Xu, Wen Li, and Lixin Duan.
\newblock Harmonious teacher for cross-domain object detection.
\newblock In \emph{Proceedings of the IEEE/CVF conference on computer vision and pattern recognition}, pages 23829--23838, 2023.

\bibitem[Devaguptapu et~al.(2019)Devaguptapu, Akolekar, M~Sharma, and N~Balasubramanian]{devaguptapu2019borrow}
Chaitanya Devaguptapu, Ninad Akolekar, Manuj M~Sharma, and Vineeth N~Balasubramanian.
\newblock Borrow from anywhere: Pseudo multi-modal object detection in thermal imagery.
\newblock In \emph{Proceedings of the IEEE/CVF Conference on Computer Vision and Pattern Recognition Workshops}, pages 0--0, 2019.

\bibitem[Ding et~al.(2021)Ding, Xue, Xia, Bai, Yang, Yang, Belongie, Luo, Datcu, Pelillo, et~al.]{ding2021object}
Jian Ding, Nan Xue, Gui-Song Xia, Xiang Bai, Wen Yang, Michael~Ying Yang, Serge Belongie, Jiebo Luo, Mihai Datcu, Marcello Pelillo, et~al.
\newblock Object detection in aerial images: A large-scale benchmark and challenges.
\newblock \emph{IEEE transactions on pattern analysis and machine intelligence}, 44\penalty0 (11):\penalty0 7778--7796, 2021.

\bibitem[Do et~al.(2024)Do, Kim, Na, Kim, Lee, Cho, and Hwang]{do2024d3t}
Dinh~Phat Do, Taehoon Kim, Jaemin Na, Jiwon Kim, Keonho Lee, Kyunghwan Cho, and Wonjun Hwang.
\newblock D3t: Distinctive dual-domain teacher zigzagging across rgb-thermal gap for domain-adaptive object detection.
\newblock In \emph{Proceedings of the IEEE/CVF Conference on Computer Vision and Pattern Recognition}, pages 23313--23322, 2024.

\bibitem[Du et~al.(2018)Du, Qi, Yu, Yang, Duan, Li, Zhang, Huang, and Tian]{du2018unmanned}
Dawei Du, Yuankai Qi, Hongyang Yu, Yifan Yang, Kaiwen Duan, Guorong Li, Weigang Zhang, Qingming Huang, and Qi Tian.
\newblock The unmanned aerial vehicle benchmark: Object detection and tracking.
\newblock In \emph{Proceedings of the European conference on computer vision (ECCV)}, pages 370--386, 2018.

\bibitem[Franchi et~al.(2024)Franchi, Hariat, Yu, Belkhir, Manzanera, and Filliat]{franchi2024infraparis}
Gianni Franchi, Marwane Hariat, Xuanlong Yu, Nacim Belkhir, Antoine Manzanera, and David Filliat.
\newblock Infraparis: A multi-modal and multi-task autonomous driving dataset.
\newblock In \emph{Proceedings of the IEEE/CVF Winter Conference on Applications of Computer Vision}, pages 2973--2983, 2024.

\bibitem[Gan et~al.(2023)Gan, Lee, and Chung]{gan2023unsupervised}
Lu Gan, Connor Lee, and Soon-Jo Chung.
\newblock Unsupervised rgb-to-thermal domain adaptation via multi-domain attention network.
\newblock In \emph{2023 IEEE International Conference on Robotics and Automation (ICRA)}, pages 6014--6020. IEEE, 2023.

\bibitem[Gao et~al.(2021)Gao, Gao, Su, Liu, Fang, Wang, and Zhang]{gao2021stereo}
Zeren Gao, Yue Gao, Yong Su, Yang Liu, Zheng Fang, Yaru Wang, and Qingchuan Zhang.
\newblock Stereo camera calibration for large field of view digital image correlation using zoom lens.
\newblock \emph{Measurement}, 185:\penalty0 109999, 2021.

\bibitem[Gong et~al.(2024)Gong, Li, Chen, and Jiang]{gong2024exploring}
Yunpeng Gong, Jiaquan Li, Lifei Chen, and Min Jiang.
\newblock Exploring color invariance through image-level ensemble learning.
\newblock \emph{arXiv preprint arXiv:2401.10512}, 2024.

\bibitem[Gonz{\'a}lez et~al.(2016)Gonz{\'a}lez, Fang, Socarras, Serrat, V{\'a}zquez, Xu, and L{\'o}pez]{gonzalez2016pedestrian}
Alejandro Gonz{\'a}lez, Zhijie Fang, Yainuvis Socarras, Joan Serrat, David V{\'a}zquez, Jiaolong Xu, and Antonio~M L{\'o}pez.
\newblock Pedestrian detection at day/night time with visible and fir cameras: A comparison.
\newblock \emph{Sensors}, 16\penalty0 (6):\penalty0 820, 2016.

\bibitem[Ha et~al.(2017)Ha, Watanabe, Karasawa, Ushiku, and Harada]{ha2017mfnet}
Qishen Ha, Kohei Watanabe, Takumi Karasawa, Yoshitaka Ushiku, and Tatsuya Harada.
\newblock Mfnet: Towards real-time semantic segmentation for autonomous vehicles with multi-spectral scenes.
\newblock In \emph{2017 IEEE/RSJ International Conference on Intelligent Robots and Systems (IROS)}, pages 5108--5115. IEEE, 2017.

\bibitem[Han et~al.(2021)Han, Ding, Xue, and Xia]{han2021redet}
Jiaming Han, Jian Ding, Nan Xue, and Gui-Song Xia.
\newblock Redet: A rotation-equivariant detector for aerial object detection.
\newblock In \emph{Proceedings of the IEEE/CVF conference on computer vision and pattern recognition}, pages 2786--2795, 2021.

\bibitem[Hou et~al.(2020)Hou, Zheng, and Gould]{hou2020learning}
Yunzhong Hou, Liang Zheng, and Stephen Gould.
\newblock Learning to structure an image with few colors.
\newblock In \emph{Proceedings of the ieee/cvf conference on computer vision and pattern recognition}, pages 10116--10125, 2020.

\bibitem[Hsu et~al.(2020)Hsu, Tsai, Lin, and Yang]{hsu2020every}
Cheng-Chun Hsu, Yi-Hsuan Tsai, Yen-Yu Lin, and Ming-Hsuan Yang.
\newblock Every pixel matters: Center-aware feature alignment for domain adaptive object detector.
\newblock In \emph{Computer Vision--ECCV 2020: 16th European Conference, Glasgow, UK, August 23--28, 2020, Proceedings, Part IX 16}, pages 733--748. Springer, 2020.

\bibitem[Hwang et~al.(2015)Hwang, Park, Kim, Choi, and Kweon]{hwang2015multispectral}
Soonmin Hwang, Jaesik Park, Namil Kim, Yukyung Choi, and In~So Kweon.
\newblock Multispectral pedestrian detection: Benchmark dataset and baselines.
\newblock In \emph{Proceedings of IEEE Conference on Computer Vision and Pattern Recognition (CVPR)}, 2015.

\bibitem[Inoue et~al.(2018)Inoue, Furuta, Yamasaki, and Aizawa]{inoue2018cross}
Naoto Inoue, Ryosuke Furuta, Toshihiko Yamasaki, and Kiyoharu Aizawa.
\newblock Cross-domain weakly-supervised object detection through progressive domain adaptation.
\newblock In \emph{Proceedings of the IEEE conference on computer vision and pattern recognition}, pages 5001--5009, 2018.

\bibitem[Jia et~al.(2021)Jia, Zhu, Li, Tang, and Zhou]{jia2021llvip}
Xinyu Jia, Chuang Zhu, Minzhen Li, Wenqi Tang, and Wenli Zhou.
\newblock Llvip: A visible-infrared paired dataset for low-light vision.
\newblock In \emph{Proceedings of the IEEE/CVF international conference on computer vision}, pages 3496--3504, 2021.

\bibitem[Jocher et~al.(2023{\natexlab{a}})Jocher, Chaurasia, and Qiu]{Jocher_Ultralytics_YOLO_2023}
Glenn Jocher, Ayush Chaurasia, and Jing Qiu.
\newblock {Ultralytics YOLO}, 2023{\natexlab{a}}.

\bibitem[Jocher et~al.(2023{\natexlab{b}})Jocher, Chaurasia, and Qiu]{yolov8_ultralytics}
Glenn Jocher, Ayush Chaurasia, and Jing Qiu.
\newblock Ultralytics yolov8, 2023{\natexlab{b}}.

\bibitem[John et~al.(2024)John, Harikumar, Senthilnath, and Sundaram]{john2024efficient}
Josy John, K Harikumar, J Senthilnath, and Suresh Sundaram.
\newblock An efficient approach with dynamic multiswarm of uavs for forest firefighting.
\newblock \emph{IEEE Transactions on Systems, Man, and Cybernetics: Systems}, 54\penalty0 (5):\penalty0 2860--2871, 2024.

\bibitem[Kim et~al.(2019)Kim, Jeong, Kim, Choi, and Kim]{kim2019diversify}
Taekyung Kim, Minki Jeong, Seunghyeon Kim, Seokeon Choi, and Changick Kim.
\newblock Diversify and match: A domain adaptive representation learning paradigm for object detection.
\newblock In \emph{Proceedings of the IEEE/CVF Conference on Computer Vision and Pattern Recognition}, pages 12456--12465, 2019.

\bibitem[Kirillov et~al.(2023)Kirillov, Mintun, Ravi, Mao, Rolland, Gustafson, Xiao, Whitehead, Berg, Lo, et~al.]{kirillov2023segment}
Alexander Kirillov, Eric Mintun, Nikhila Ravi, Hanzi Mao, Chloe Rolland, Laura Gustafson, Tete Xiao, Spencer Whitehead, Alexander~C Berg, Wan-Yen Lo, et~al.
\newblock Segment anything.
\newblock In \emph{Proceedings of the IEEE/CVF International Conference on Computer Vision}, pages 4015--4026, 2023.

\bibitem[K{\"u}t{\"u}k and Algan(2022)]{kutuk2022semantic}
Z{\"u}lfiye K{\"u}t{\"u}k and G{\"o}rkem Algan.
\newblock Semantic segmentation for thermal images: A comparative survey.
\newblock In \emph{Proceedings of the IEEE/CVF Conference on Computer Vision and Pattern Recognition}, pages 286--295, 2022.

\bibitem[Lee et~al.(2024)Lee, Chen, Dam, Ferdaus, Poenar, and Duong]{lee2024dehazing}
Gao~Yu Lee, Jinkuan Chen, Tanmoy Dam, Md~Meftahul Ferdaus, Daniel~Puiu Poenar, and Vu~N Duong.
\newblock Dehazing remote sensing and uav imagery: A review of deep learning, prior-based, and hybrid approaches.
\newblock \emph{arXiv preprint arXiv:2405.07520}, 2024.

\bibitem[Li et~al.(2022)Li, Dai, Ma, Liu, Chen, Wu, He, Kitani, and Vajda]{li2022cross}
Yu-Jhe Li, Xiaoliang Dai, Chih-Yao Ma, Yen-Cheng Liu, Kan Chen, Bichen Wu, Zijian He, Kris Kitani, and Peter Vajda.
\newblock Cross-domain adaptive teacher for object detection.
\newblock In \emph{IEEE Conference on Computer Vision and Pattern Recognition (CVPR)}, 2022.

\bibitem[Liu et~al.(2022)Liu, Fan, Huang, Wu, Liu, Zhong, and Luo]{liu2022target}
Jinyuan Liu, Xin Fan, Zhanbo Huang, Guanyao Wu, Risheng Liu, Wei Zhong, and Zhongxuan Luo.
\newblock Target-aware dual adversarial learning and a multi-scenario multi-modality benchmark to fuse infrared and visible for object detection.
\newblock In \emph{Proceedings of the IEEE/CVF conference on computer vision and pattern recognition}, pages 5802--5811, 2022.

\bibitem[Liu et~al.(2021)Liu, Ma, He, Kuo, Chen, Zhang, Wu, Kira, and Vajda]{liu2021unbiased}
Yen-Cheng Liu, Chih-Yao Ma, Zijian He, Chia-Wen Kuo, Kan Chen, Peizhao Zhang, Bichen Wu, Zsolt Kira, and Peter Vajda.
\newblock Unbiased teacher for semi-supervised object detection.
\newblock \emph{arXiv preprint arXiv:2102.09480}, 2021.

\bibitem[Lowe(2004)]{lowe2004distinctive}
David~G Lowe.
\newblock Distinctive image features from scale-invariant keypoints.
\newblock \emph{International journal of computer vision}, 60:\penalty0 91--110, 2004.

\bibitem[Ma et~al.(2024{\natexlab{a}})Ma, Su, Li, and Xian]{ma2024aerialirgan}
Decao Ma, Juan Su, Shaopeng Li, and Yong Xian.
\newblock Aerialirgan: unpaired aerial visible-to-infrared image translation with dual-encoder structure.
\newblock \emph{Scientific Reports}, 14\penalty0 (1):\penalty0 22105, 2024{\natexlab{a}}.

\bibitem[Ma et~al.(2024{\natexlab{b}})Ma, Chai, Jin, and Yan]{ma2024hierarchical}
You Ma, Lin Chai, Lizuo Jin, and Jun Yan.
\newblock Hierarchical alignment network for domain adaptive object detection in aerial images.
\newblock \emph{ISPRS Journal of Photogrammetry and Remote Sensing}, 208:\penalty0 39--52, 2024{\natexlab{b}}.

\bibitem[Razakarivony and Jurie(2016)]{razakarivony2016vehicle}
Sebastien Razakarivony and Frederic Jurie.
\newblock Vehicle detection in aerial imagery: A small target detection benchmark.
\newblock \emph{Journal of Visual Communication and Image Representation}, 34:\penalty0 187--203, 2016.

\bibitem[Ren et~al.(2015)Ren, He, Girshick, and Sun]{NIPS2015_14bfa6bb}
Shaoqing Ren, Kaiming He, Ross Girshick, and Jian Sun.
\newblock Faster r-cnn: Towards real-time object detection with region proposal networks.
\newblock In \emph{Advances in Neural Information Processing Systems}. Curran Associates, Inc., 2015.

\bibitem[Sarker et~al.(2024)Sarker, Zhao, and Uddin]{sarker2024transformer}
Prodip~Kumar Sarker, Qingjie Zhao, and Md~Kamal Uddin.
\newblock Transformer-based person re-identification: A comprehensive review.
\newblock \emph{IEEE Transactions on Intelligent Vehicles}, 9\penalty0 (7):\penalty0 5222--5239, 2024.

\bibitem[Senthilnath et~al.(2024)Senthilnath, Harikumar, and Sundaram]{senthilnath2024metacognitive}
J Senthilnath, K Harikumar, and Suresh Sundaram.
\newblock Metacognitive decision-making framework for multi-uav target search without communication.
\newblock \emph{IEEE Transactions on Systems, Man, and Cybernetics: Systems}, 54\penalty0 (5):\penalty0 3195--3206, 2024.

\bibitem[Sikdar et~al.(2022)Sikdar, Udupa, Sundaram, and Sundararajan]{sikdar2022fully}
Aniruddh Sikdar, Sumanth Udupa, Suresh Sundaram, and Narasimhan Sundararajan.
\newblock Fully complex-valued fully convolutional multi-feature fusion network (fc 2 mfn) for building segmentation of insar images.
\newblock In \emph{2022 IEEE Symposium Series on Computational Intelligence (SSCI)}, pages 581--587. IEEE, 2022.

\bibitem[Sikdar et~al.(2023{\natexlab{a}})Sikdar, Teotia, and Sundaram]{sikdar2023contrastive}
Aniruddh Sikdar, Jayant Teotia, and Suresh Sundaram.
\newblock Contrastive learning-based spectral knowledge distillation for multi-modality and missing modality scenarios in semantic segmentation.
\newblock \emph{arXiv preprint arXiv:2312.02240}, 2023{\natexlab{a}}.

\bibitem[Sikdar et~al.(2023{\natexlab{b}})Sikdar, Udupa, Gurunath, and Sundaram]{sikdar2023deepmao}
Aniruddh Sikdar, Sumanth Udupa, Prajwal Gurunath, and Suresh Sundaram.
\newblock Deepmao: Deep multi-scale aware overcomplete network for building segmentation in satellite imagery.
\newblock In \emph{Proceedings of the IEEE/CVF Conference on Computer Vision and Pattern Recognition}, pages 487--496, 2023{\natexlab{b}}.

\bibitem[Sikdar et~al.(2024{\natexlab{a}})Sikdar, Saadiyean, Anand, and Sundaram]{sikdar2024ssl}
Aniruddh Sikdar, Qiranul Saadiyean, Prahlad Anand, and Suresh Sundaram.
\newblock Ssl-rgb2ir: Semi-supervised rgb-to-ir image-to-image translation for enhancing visual task training in semantic segmentation and object detection.
\newblock In \emph{2024 IEEE/RSJ International Conference on Intelligent Robots and Systems (IROS)}, pages 5017--5023. IEEE, 2024{\natexlab{a}}.

\bibitem[Sikdar et~al.(2024{\natexlab{b}})Sikdar, Teotia, and Sundaram]{sikdar2024skd}
Aniruddh Sikdar, Jayant Teotia, and Suresh Sundaram.
\newblock Skd-net: Spectral-based knowledge distillation in low-light thermal imagery for robotic perception.
\newblock In \emph{2024 IEEE International Conference on Robotics and Automation (ICRA)}, pages 9041--9047. IEEE, 2024{\natexlab{b}}.

\bibitem[Sikdar et~al.(2025)Sikdar, Teotia, and Sundaram]{Sikdar_Teotia_Sundaram_2025}
Aniruddh Sikdar, Jayant Teotia, and Suresh Sundaram.
\newblock Ogp-net: Optical guidance meets pixel-level contrastive distillation for robust multi-modal and missing modality segmentation.
\newblock \emph{Proceedings of the AAAI Conference on Artificial Intelligence}, 39\penalty0 (7):\penalty0 6922--6930, 2025.

\bibitem[Song et~al.(2024)Song, Xue, Wen, Ji, Yan, and Meng]{song2024misaligned}
Kechen Song, Xiaotong Xue, Hongwei Wen, Yingying Ji, Yunhui Yan, and Qinggang Meng.
\newblock Misaligned visible-thermal object detection: a drone-based benchmark and baseline.
\newblock \emph{IEEE Transactions on Intelligent Vehicles}, 2024.

\bibitem[source FLIR dataset~available online()]{fliresh}
Open source FLIR dataset~available online.
\newblock Free teledyne flir thermal dataset for algorithm training.
\newblock \emph{https://www.flir.com/oem/adas/adas-dataset-form/}.

\bibitem[Sun et~al.(2022)Sun, Cao, Zhu, and Hu]{sun2020drone}
Yiming Sun, Bing Cao, Pengfei Zhu, and Qinghua Hu.
\newblock Drone-based rgb-infrared cross-modality vehicle detection via uncertainty-aware learning.
\newblock \emph{IEEE Transactions on Circuits and Systems for Video Technology}, pages 1--1, 2022.

\bibitem[Suo et~al.(2023)Suo, Wang, Zhang, Chen, Zhou, and Shi]{suo2023hit}
Jiashun Suo, Tianyi Wang, Xingzhou Zhang, Haiyang Chen, Wei Zhou, and Weisong Shi.
\newblock Hit-uav: A high-altitude infrared thermal dataset for unmanned aerial vehicle-based object detection.
\newblock \emph{Scientific Data}, 10\penalty0 (1):\penalty0 227, 2023.

\bibitem[Tang et~al.(2022)Tang, Yuan, Zhang, Jiang, and Ma]{tang2022piafusion}
Linfeng Tang, Jiteng Yuan, Hao Zhang, Xingyu Jiang, and Jiayi Ma.
\newblock Piafusion: A progressive infrared and visible image fusion network based on illumination aware.
\newblock \emph{Information Fusion}, 83:\penalty0 79--92, 2022.

\bibitem[Tarvainen and Valpola(2017)]{tarvainen2017mean}
Antti Tarvainen and Harri Valpola.
\newblock Mean teachers are better role models: Weight-averaged consistency targets improve semi-supervised deep learning results.
\newblock \emph{Advances in neural information processing systems}, 30, 2017.

\bibitem[Tennant(2011)]{tennant2011limits}
WE Tennant.
\newblock Limits of infrared imaging.
\newblock \emph{International Journal of High Speed Electronics and Systems}, 20\penalty0 (03):\penalty0 529--539, 2011.

\bibitem[Udupa et~al.(2024)Udupa, Gurunath, Sikdar, and Sundaram]{udupa2024mrfp}
Sumanth Udupa, Prajwal Gurunath, Aniruddh Sikdar, and Suresh Sundaram.
\newblock Mrfp: Learning generalizable semantic segmentation from sim-2-real with multi-resolution feature perturbation.
\newblock In \emph{Proceedings of the IEEE/CVF Conference on Computer Vision and Pattern Recognition}, pages 5904--5914, 2024.

\bibitem[Ustun et~al.(2023)Ustun, Kaya, Ayerden, and Altinel]{ustun2023spectral}
Berkcan Ustun, Ahmet~Kagan Kaya, Ezgi~Cakir Ayerden, and Fazil Altinel.
\newblock Spectral transfer guided active domain adaptation for thermal imagery.
\newblock In \emph{Proceedings of the IEEE/CVF Conference on Computer Vision and Pattern Recognition}, pages 449--458, 2023.

\bibitem[Valada et~al.(2020)Valada, Mohan, and Burgard]{valada2020self}
Abhinav Valada, Rohit Mohan, and Wolfram Burgard.
\newblock Self-supervised model adaptation for multimodal semantic segmentation.
\newblock \emph{International Journal of Computer Vision}, 128\penalty0 (5):\penalty0 1239--1285, 2020.

\bibitem[Wang et~al.(2024)Wang, Zhang, Du, Xu, Liu, Tao, and Zhang]{wang2024samrs}
Di Wang, Jing Zhang, Bo Du, Minqiang Xu, Lin Liu, Dacheng Tao, and Liangpei Zhang.
\newblock Samrs: Scaling-up remote sensing segmentation dataset with segment anything model.
\newblock \emph{Advances in Neural Information Processing Systems}, 36, 2024.

\bibitem[Wang(2023)]{X-AnyLabeling}
Wei Wang.
\newblock Advanced auto labeling solution with added features.
\newblock \url{https://github.com/CVHub520/X-AnyLabeling}, 2023.

\bibitem[Wu et~al.(2017)Wu, Zheng, Yu, Gong, and Lai]{wu2017rgb}
Ancong Wu, Wei-Shi Zheng, Hong-Xing Yu, Shaogang Gong, and Jianhuang Lai.
\newblock Rgb-infrared cross-modality person re-identification.
\newblock In \emph{Proceedings of the IEEE international conference on computer vision}, pages 5380--5389, 2017.

\bibitem[Xia et~al.(2018)Xia, Bai, Ding, Zhu, Belongie, Luo, Datcu, Pelillo, and Zhang]{xia2018dota}
Gui-Song Xia, Xiang Bai, Jian Ding, Zhen Zhu, Serge Belongie, Jiebo Luo, Mihai Datcu, Marcello Pelillo, and Liangpei Zhang.
\newblock Dota: A large-scale dataset for object detection in aerial images.
\newblock In \emph{Proceedings of the IEEE conference on computer vision and pattern recognition}, pages 3974--3983, 2018.

\bibitem[Xie et~al.(2021)Xie, Cheng, Wang, Yao, and Han]{xie2021oriented}
Xingxing Xie, Gong Cheng, Jiabao Wang, Xiwen Yao, and Junwei Han.
\newblock Oriented r-cnn for object detection.
\newblock In \emph{Proceedings of the IEEE/CVF international conference on computer vision}, pages 3520--3529, 2021.

\bibitem[Zeng et~al.(2024)Zeng, Gu, Qin, Jia, Deng, Xu, and Tian]{zeng2024unsupervised}
Junying Zeng, Yajin Gu, Chuanbo Qin, Xudong Jia, Senyao Deng, Jiahua Xu, and Huiming Tian.
\newblock Unsupervised domain adaptation for remote sensing semantic segmentation with the 2d discrete wavelet transform.
\newblock \emph{Scientific Reports}, 14\penalty0 (1):\penalty0 23552, 2024.

\bibitem[Zhang et~al.(2020)Zhang, Fromont, Lefevre, and Avignon]{zhang2020multispectral}
Heng Zhang, Elisa Fromont, S{\'e}bastien Lefevre, and Bruno Avignon.
\newblock Multispectral fusion for object detection with cyclic fuse-and-refine blocks.
\newblock In \emph{2020 IEEE International conference on image processing (ICIP)}, pages 276--280. IEEE, 2020.

\bibitem[Zhang et~al.(2022)Zhang, Lai, Liu, Huang, and Han]{zhang2022fmcnet}
Qiang Zhang, Changzhou Lai, Jianan Liu, Nianchang Huang, and Jungong Han.
\newblock Fmcnet: Feature-level modality compensation for visible-infrared person re-identification.
\newblock In \emph{Proceedings of the IEEE/CVF conference on computer vision and pattern recognition}, pages 7349--7358, 2022.

\bibitem[Zhao et~al.(2021)Zhao, Liu, Chu, Lu, and Yu]{zhao2021joint}
Zhiwei Zhao, Bin Liu, Qi Chu, Yan Lu, and Nenghai Yu.
\newblock Joint color-irrelevant consistency learning and identity-aware modality adaptation for visible-infrared cross modality person re-identification.
\newblock In \emph{Proceedings of the AAAI conference on artificial intelligence}, pages 3520--3528, 2021.

\bibitem[Zhu et~al.(2021)Zhu, Wen, Du, Bian, Fan, Hu, and Ling]{zhu2021detection}
Pengfei Zhu, Longyin Wen, Dawei Du, Xiao Bian, Heng Fan, Qinghua Hu, and Haibin Ling.
\newblock Detection and tracking meet drones challenge.
\newblock \emph{IEEE Transactions on Pattern Analysis and Machine Intelligence}, 44\penalty0 (11):\penalty0 7380--7399, 2021.

\end{thebibliography}
